\documentclass{article}


\usepackage{microtype}
\usepackage{graphicx}
\usepackage{subcaption}
\usepackage{booktabs} % for professional tables

\usepackage[preprint]{icml2026}

\usepackage{algorithm}
\usepackage{algpseudocode}
\usepackage{hyperref}

\usepackage{amsmath}
\usepackage{amssymb}
\usepackage{mathtools}
\usepackage{amsthm}

\usepackage[capitalize,noabbrev]{cleveref}

\theoremstyle{plain}

\theoremstyle{definition}

\theoremstyle{remark}

\usepackage[textsize=tiny]{todonotes}

\icmltitlerunning{DP-OPD: Differentially Private On-Policy Distillation for Language Models}

\begin{document}

\twocolumn[
\icmltitle{DP-OPD: Differentially Private On-Policy Distillation for Language Models}

\begin{icmlauthorlist}
  \icmlauthor{Fatemeh Khadem}{scu}
  \icmlauthor{Sajad Mousavi}{ind}
  \icmlauthor{Yi Fang}{scu}
  \icmlauthor{Yuhong Liu}{scu}
\end{icmlauthorlist}

\icmlaffiliation{scu}{Santa Clara University, Santa Clara, California, USA}
\icmlaffiliation{ind}{Independent Researcher}

\icmlcorrespondingauthor{Fatemeh Khadem}{fkhadem@scu.edu}
\icmlcorrespondingauthor{Sajad Mousavi}{smousavi71@gmail.com}
\icmlcorrespondingauthor{Yi Fang}{yfang@scu.edu}
\icmlcorrespondingauthor{Yuhong Liu}{yhliu@scu.edu}

\icmlkeywords{Differential Privacy, Knowledge Distillation, On-Policy Distillation, Language Models}

\vskip 0.3in
]

\printAffiliationsAndNotice{}
\begin{abstract}
Large language models (LLMs) are increasingly adapted to proprietary and domain-specific corpora that contain sensitive information, creating a tension between formal privacy guarantees and efficient deployment through model compression. Differential privacy (DP), typically enforced via DP-SGD, provides record-level protection but often incurs substantial utility loss in autoregressive generation, where optimization noise can amplify exposure bias and compounding errors along long rollouts. Existing approaches to private distillation either apply DP-SGD to both teacher and student, worsening computation and the privacy--utility tradeoff, or rely on DP synthetic text generation from a DP-trained teacher, avoiding DP on the student at the cost of DP-optimizing a large teacher and introducing an offline generation pipeline. We propose \textbf{Differentially Private On-Policy Distillation (DP-OPD)}, a synthesis-free framework that enforces privacy solely through DP-SGD on the student while leveraging a frozen teacher to provide dense token-level targets on \emph{student-generated} trajectories. DP-OPD instantiates this idea via \emph{private generalized knowledge distillation} on continuation tokens. Under a strict privacy budget ($\varepsilon=2.0$), DP-OPD improves perplexity over DP fine-tuning and off-policy DP distillation, and outperforms synthesis-based DP distillation (Yelp: 44.15$\rightarrow$41.68; BigPatent: 32.43$\rightarrow$30.63), while substantially simplifying the training pipeline. In particular, \textbf{DP-OPD collapses private compression into a single DP student-training loop} by eliminating DP teacher training and offline synthetic text generation.
Code will be released upon publication at \url{https://github.com/khademfatemeh/dp_opd}.
\end{abstract}

\section{Introduction}

Large language models (LLMs) are increasingly adapted to domain- and organization-specific corpora that contain sensitive information.
This creates a tension between two practical requirements: (i) \emph{formal privacy guarantees} for training records and (ii) \emph{efficient deployment}
through model compression.
In practice, many deployments face strict constraints on latency, memory footprint, and serving cost (or require on-device inference), motivating compression of large teachers into smaller students.

Differential privacy (DP) provides a principled protection against information leakage from individual training records~\cite{dwork2006calibrating},
but in modern LMs it often comes with substantial utility loss because DP optimization injects noise into the learning dynamics~\cite{abadi2016deep}
and can amplify distributional errors in autoregressive generation.
At the same time, knowledge distillation (KD) is a standard technique for compressing LMs by transferring behavior from a large teacher to a smaller student,
yet conventional KD for sequence models is typically \emph{off-policy}: it trains on fixed, teacher-forced trajectories that differ from the student’s
own generation-time state distribution in autoregressive \emph{token} generation (exposure bias/compounding errors)~\cite{bengio2015scheduled,ranzato2016sequence}.

\begin{figure*}[t]
  \centering
  \begin{subfigure}{0.48\linewidth}
    \centering
    \includegraphics[width=\linewidth]{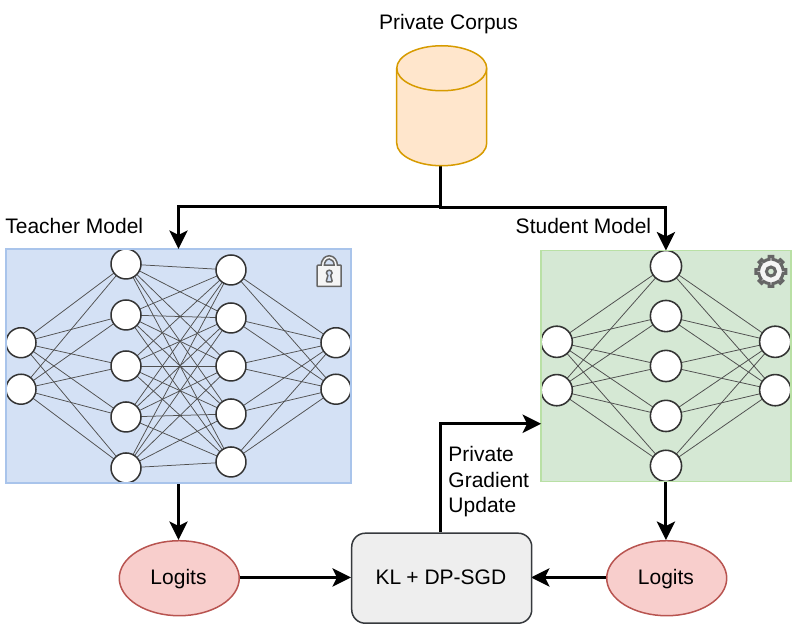}
    \caption{DP off-policy KD (teacher-forced / fixed trajectories).}
    \label{fig:dp_off_policy_kd}
  \end{subfigure}\hfill
  \begin{subfigure}{0.48\linewidth}
    \centering
    \includegraphics[width=\linewidth]{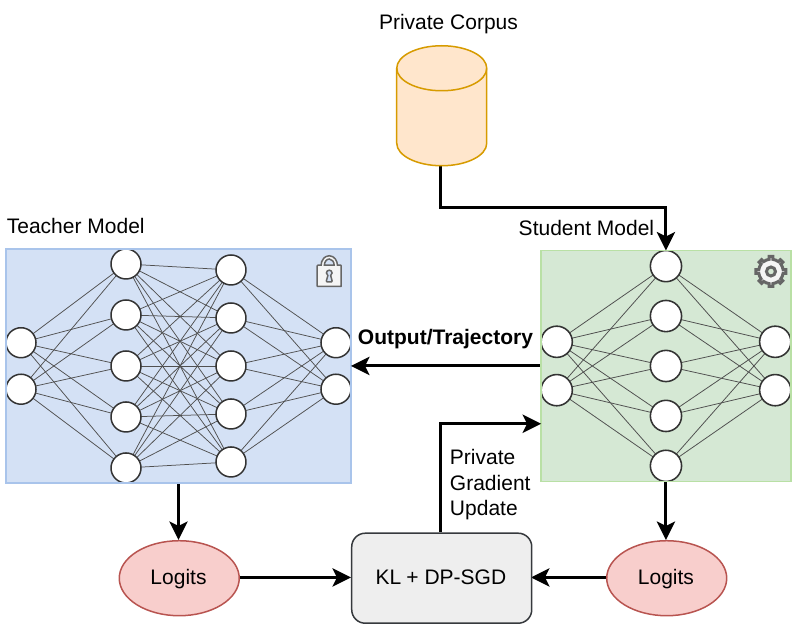}
    \caption{DP on-policy KD (DP-OPD): distill on student rollouts.}
    \label{fig:dp_on_policy_kd}
  \end{subfigure}
  \caption{Contrasting DP distillation paradigms. In DP-OPD, teacher targets are evaluated on student-generated trajectories, while privacy is enforced solely via DP-SGD on student updates.}
  \label{fig:dp_kd_contrast}
\end{figure*}

Figure~\ref{fig:dp_kd_contrast} contrasts the standard \emph{DP off-policy} distillation setup---where distillation targets are computed on fixed or
teacher-forced trajectories---with \emph{DP on-policy} distillation, where the teacher provides token-level targets on student-generated rollouts and
privacy is enforced solely through differentially private stochastic gradient descent (DP-SGD)~\cite{abadi2016deep} on the student updates.
Note that the student is trained on the private corpus (as in standard DP training), and DP is needed precisely because it bounds the influence of any single training record on the released student parameters.
The teacher is assumed public (e.g., pretrained) or trained independently of the private corpus, and is used only to provide targets during training.

A straightforward way to combine DP and KD is to train (or fine-tune) the teacher with DP-SGD and then apply DP-SGD again when training the student
during distillation; however, this \emph{dual-DP} (teacher DP-SGD + student DP-SGD) approach incurs substantial additional computation and often worsens the privacy--utility tradeoff
in strict privacy regimes~\cite{mireshghallah2022differentially}.

Recent work has begun to connect DP and distillation for LMs by using \emph{DP synthetic text}.
\textsc{DistilDP}~\cite{distildp} trains a differentially private teacher with DP-SGD, generates a large synthetic corpus from that DP teacher
(a DP post-processing step), and then distills a student on the synthetic data using both hard tokens and soft teacher distributions.
This pipeline is appealing because it avoids DP-SGD on the student; however, it shifts the burden to (i) DP training of a large teacher,
and (ii) an offline synthetic data stage that introduces additional algorithmic and engineering choices (decoding strategy, synthetic set size,
and dataset-specific prompt/conditioning design).

We propose \textbf{Differentially Private On-Policy Distillation (DP-OPD)}, a synthesis-free alternative that applies DP \emph{only} to the student updates
while still leveraging a strong teacher signal.
DP-OPD is motivated by \emph{on-policy distillation}~\cite{agarwal2024opd}, which addresses train--test mismatch in autoregressive models by training the
student on its \emph{self-generated} continuations and querying a teacher to provide dense token-level targets on those trajectories.
This principle is especially relevant under DP: when optimization noise perturbs the student, small local mistakes can compound along a rollout,
so matching the teacher \emph{on the states the student actually visits} becomes critical.

DP-OPD instantiates this idea with a \emph{generalized} distillation objective (Generalized KD; GKD) that flexibly interpolates between forward-KL,
reverse-KL, and JSD-style divergences, enabling robust transfer even when the student lacks capacity to exactly match the teacher distribution~\cite{agarwal2024opd}.
In conditional generation settings, we encode structured attributes as textual \emph{control codes} prepended to the prompt; this affects only the input representation (a standardized prefix template) and does not change the DP-OPD optimization procedure.
Using a uniform control-code interface enables consistent evaluation across domains by expressing dataset-specific conditioning (e.g., labels/metadata) in the same textual format~\cite{distildp}.
Across short- and long-form text regimes, DP-OPD yields an improved privacy--utility tradeoff (measured by test perplexity) while avoiding DP teacher training and offline synthetic text generation.

\paragraph{Contributions.}
Together, our contributions improve utility under strict privacy, eliminate the need to privately train a large teacher (i.e. applying DP on it) and remove the offline synthetic generation stage, yielding a simpler and more compute-efficient private compression pipeline.
We make three main contributions:
(1) We introduce \textbf{DP-OPD}, a synthesis-free differentially private distillation framework for autoregressive LMs that performs \emph{on-policy}
teacher--student alignment while privatizing only the student updates.
(2) We instantiate DP-OPD with \textbf{private generalized KD} on continuation tokens, connecting on-policy distillation~\cite{agarwal2024opd} to
standard DP-SGD accounting~\cite{abadi2016deep}.
(3) We provide a practical and general training recipe that eliminates DP training of the teacher and offline synthetic text generation, substantially
simplifying private compression while retaining dense teacher guidance on student rollouts, and achieving a stronger privacy--utility tradeoff than standard DP off-policy knowledge distillation and synthesis-based DP distillation in strict privacy regimes.

\section{Related Work}

\paragraph{Differential privacy for training language models.}
DP provides record-level protection by bounding the influence of any single training example on the released model parameters.
In deep learning, the dominant approach is DP-SGD, which clips per-example gradients and adds Gaussian noise, with privacy tracked by composition
accounting under subsampling~\cite{abadi2016deep,dwork2006calibrating}.
Applying DP-SGD to NLP models has historically been challenging due to utility degradation and computational overhead, but recent studies show that
large pretrained models can substantially improve DP fine-tuning outcomes under the same privacy budget~\cite{li2021large}.
Beyond training-time mechanisms, complementary directions include DP decoding-time perturbations for already-trained LMs~\cite{majmudar2022dpdecoding}
and DP generative models for releasing privacy-preserving text corpora~\cite{mattern2022dplm}.

\paragraph{Private learning via teachers, distillation, and synthetic text.}
A classical privacy-preserving paradigm is to avoid releasing models trained directly on private data and instead transfer knowledge through privatized
signals from teacher models, e.g., PATE-style teacher ensembles~\cite{papernot2016pate}.
For LMs, recent work revisits this idea through \emph{synthetic text generation} from DP models: once a generator is DP, sampling from it is
post-processing and thus preserves privacy.
\textsc{DistilDP}~\cite{distildp} operationalizes this pipeline for autoregressive LMs by training a DP teacher, generating DP synthetic text,
and then distilling a student on that synthetic corpus. Related efforts also explore private synthetic text generation with LMs and decoding strategies
that reduce privacy leakage~\cite{kurakin2023privateSynth,majmudar2022dpdecoding}.
These methods highlight a core tradeoff: moving DP to a teacher/synthetic stage can reduce DP training burden on the student, but it introduces an
additional generation pipeline and typically requires DP optimization on large models.

\paragraph{Knowledge distillation for sequence models and exposure bias.}
Standard KD transfers teacher behavior by aligning student predictions with teacher distributions on a fixed dataset.
For autoregressive generation, however, training typically conditions on ground-truth prefixes, whereas inference conditions on the model’s own prefixes,
creating \emph{exposure bias} and compounding errors~\cite{bengio2015scheduled,ranzato2016sequence}.
This mismatch is consequential even without privacy, and it can become more pronounced under DP noise because local errors can propagate through a rollout.

\paragraph{On-policy distillation and generalized KD.}
On-policy distillation addresses exposure bias by sampling trajectories from the student and supervising the student with teacher feedback on the states
the student actually visits.
Agarwal et al.~\cite{agarwal2024opd} formalize this view and propose \emph{Generalized Knowledge Distillation (GKD)}, which allows alternative divergence
measures (including JSD-style objectives) between teacher and student distributions to accommodate limited student expressivity.
In practice, on-policy distillation has also motivated engineering solutions for tokenizer mismatch and cross-family distillation (e.g., logit alignment),
enabling teacher supervision even when vocabularies differ.

\paragraph{Positioning of DP-OPD.}
DP-OPD sits at the intersection of these lines of work: it brings on-policy distillation and GKD-style objectives into the differentially private training
regime.
In contrast to synthesis-based DP distillation (e.g., \textsc{DistilDP}~\cite{distildp}), DP-OPD removes offline synthetic generation and avoids DP training
of the teacher; privacy is enforced solely through DP-SGD on the student updates.
In contrast to DP fine-tuning without a teacher, DP-OPD provides dense token-level guidance on student rollouts, explicitly targeting the inference-time
state distribution where exposure bias and DP-induced errors manifest.

\section{Method: Differentially Private On-Policy Distillation (DP-OPD)}
\label{sec:method}

\subsection{Setup}

Let $D=\{(x_i, y_i)\}_{i=1}^{N}$ be a private dataset of text sequences. Each $x_i$ denotes a prompt derived from record $i$ and $y_i$ is an
optional reference continuation. Prompts may include dataset-specific conditioning information (\emph{control code}) such as metadata or class label
prepended to the text, as in prior work \cite{distildp}  $c_i$ derived from structured attributes (e.g., label/metadata),
and define the model input as $\tilde{x}_i = c_i \oplus x_i$. This prompt formatting is orthogonal to DP-OPD. 

We assume access to a fixed teacher language model $T$ (public / non-private, frozen) and a student model
$S_\theta$ to be trained. Our goal is to output $\theta$ such that the trained student is $(\varepsilon,\delta)$-DP with respect to $D$ while maintaining generation quality.

\subsection{On-policy rollouts}

At each training step, we sample a minibatch $\mathcal{I}$ of size $B$ using DP-compatible subsampling (sampling rate $q=B/N$).
With probability $\lambda\in[0,1]$, we perform an \emph{on-policy} step by sampling student continuations:
\begin{equation}
\hat{y}_i \sim S_\theta(\cdot \mid \tilde{x}_i), \qquad i\in\mathcal{I},
\end{equation}
truncated to at most $L$ new tokens.
Otherwise (with probability $1-\lambda$), we use the teacher-forced sequences provided by the batch (off-policy).

In both cases, the loss is computed \emph{only on continuation tokens} (prompt tokens are masked out).

\subsection{Generalized knowledge distillation loss }

Let $z_S(\cdot\mid s)$ and $z_T(\cdot\mid s)$ be student/teacher logits at state $s$, and let $\tau_d>0$ be the distillation temperature.
We define temperature-scaled distributions:
\begin{align}
p_S(\cdot\mid s)
  &= \mathrm{softmax}\!\left(z_S(\cdot\mid s)/\tau_d\right), \\
p_T(\cdot\mid s)
  &= \mathrm{softmax}\!\left(z_T(\cdot\mid s)/\tau_d\right).
\end{align}

We minimize a \emph{generalized} distillation divergence ~\cite{agarwal2024opd} on continuation tokens, parameterized by $\beta\in[0,1]$:
$\beta=0$ corresponds to forward-KL-style distillation, $\beta=1$ corresponds to reverse-KL-style distillation, and $\beta=0.5$ yields a JSD-like objective.
We denote the resulting per-example loss (on masked continuation positions) as
\begin{equation}
\ell_i(\theta) \;=\; \mathcal{L}_{\mathrm{GKD}}\!\left(p_S(\cdot\mid \tilde{x}_i,\cdot),\, p_T(\cdot\mid \tilde{x}_i,\cdot);\; \beta\right),
\label{eq:gkd}
\end{equation}
where masking ensures that only continuation tokens contribute. When hard labels are available (teacher-forced or rollout labels), they are incorporated
in $\mathcal{L}_{\mathrm{GKD}}$ as in standard generalized KD formulations.

Intuitively, standard off-policy KD matches the teacher on fixed, teacher-forced trajectories, which can differ from the states encountered at inference.
In contrast, DP-OPD evaluates the teacher on student rollouts and applies distillation on continuation tokens under the student policy.
As a result, the student is trained to agree with the teacher on the trajectories it actually visits, reducing exposure bias and mitigating
compounding errors when the model conditions on its own outputs at test time, a behavior that becomes more pronounced under DP noise.

\subsection{Differential privacy: DP-SGD on student updates}

We optimize the per-example objective $\ell_i(\theta)$ (e.g., \cref{eq:gkd}) using DP-SGD.
At each step, we sample a minibatch $\mathcal{I}$ of size $B$ with DP-compatible subsampling, compute per-example gradients
$g_i=\nabla_\theta \ell_i(\theta)$, clip them to $\ell_2$ norm $C$, and add Gaussian noise with multiplier $\sigma$:
\begin{equation}
\tilde{g}=\frac{1}{B}\sum_{i\in\mathcal{I}}\mathrm{clip}_C(g_i)
\;+\; \frac{1}{B}\mathcal{N}\!\left(0, \sigma^2 C^2 I\right),
\qquad
\theta \leftarrow \theta - \eta\,\tilde{g}.
\label{eq:dpsgd}
\end{equation}
We track the resulting privacy guarantee $(\varepsilon,\delta)$ using a standard accountant (e.g., RDP) under the chosen subsampling scheme.

\paragraph{What consumes privacy.}
The teacher $T$ is frozen and never released; it is used only to produce targets within the loss.
The only released artifact is the final student parameters $\theta$, and the privacy guarantee is governed entirely by the DP-SGD update mechanism.
In particular, teacher inference does not consume additional privacy budget because it is internal computation within the DP training pipeline,
and DP is accounted for solely through the student updates with respect to the private dataset $D$.

\begin{algorithm}[t]
\caption{\textbf{DP-OPD}: Differentially Private On-Policy Distillation}
\label{alg:dp-opd}
\begin{algorithmic}[1]
\Require Private dataset $D$; optional control codes $\{c_i\}$; student $S_\theta$; frozen teacher $T$;
on-policy probability $\lambda$; max new tokens $L$; distill temperature $\tau_d$; GKD parameter $\beta$;
DP-SGD params $(B,C,\sigma)$; steps $U$; target $\delta$.
\For{$u=1$ \textbf{to} $U$}
  \State Sample minibatch $\mathcal{I}$ of size $B$ with DP-compatible subsampling.
  \For{each $i\in\mathcal{I}$}
    \State $\tilde{x}_i \leftarrow c_i \oplus x_i$ \Comment{$c_i=\emptyset$ if unused}
  \EndFor
  \If{with probability $\lambda$} \Comment{On-policy}
    \For{each $i\in\mathcal{I}$}
      \State Sample $\hat{y}_i \sim S_\theta(\cdot\mid \tilde{x}_i)$, truncated to $L$ tokens.
      \State $s_i \leftarrow \tilde{x}_i \oplus \hat{y}_i$; build continuation mask (mask prompt positions).
    \EndFor
  \Else \Comment{Off-policy / teacher-forcing}
    \For{each $i\in\mathcal{I}$}
      \State $s_i \leftarrow \tilde{x}_i \oplus y_i$; build continuation mask (mask prompt positions).
    \EndFor
  \EndIf
  \For{each $i\in\mathcal{I}$}
    \State Compute logits $z_S(\cdot \mid s_i)$ and $z_T(\cdot \mid s_i)$ (next-token distributions).
    \State $\ell_i \leftarrow \mathcal{L}_{\mathrm{GKD}}(z_S,z_T;\beta,\tau_d)$ on continuation tokens only.
  \EndFor
  \State $\theta \leftarrow \textsc{DP-SGD-Step}(\theta,\{\ell_i\}_{i\in\mathcal{I}}, C,\sigma)$; update accountant for target $\delta$.
\EndFor
\State \Return $\theta$
\end{algorithmic}
\end{algorithm}

\paragraph{Tokenizer mismatch.}
Our main experiments use teacher and student models with the same tokenizer for simplicity. If teacher and student tokenizers differ, on-policy
distillation can be extended using logit-alignment methods such as GOLD~\cite{opd_h4_space}.

\section{Experiments}
\label{sec:experiments}

We evaluate whether \textbf{DP-OPD} improves the privacy--utility tradeoff for differentially private compression of autoregressive language models.
We compare against (i) private fine-tuning of the student without a teacher (\textbf{DP-SGD}), (ii) a two-stage off-policy private distillation baseline that applies DP-SGD to both teacher and student (\textbf{DPKD}), and (iii) synthesis-based private distillation via DP synthetic text (\textbf{\textsc{DistilDP}}).
To enable a directly comparable benchmark, we mirror \textsc{DistilDP}'s preprocessing, control-code conditioning, and evaluation protocol, and we use their reported baseline perplexities when available.

\subsection{Datasets and conditioning}
\label{sec:datasets}

\textbf{Yelp (short-form reviews).}
We use the Yelp Open Dataset.\footnote{\url{https://www.yelp.com/dataset}}
Following prior synthetic-text DP training recipes \cite{yue2023synthetic} and the \textsc{DistilDP} benchmark setup, we keep the top-10 business categories, remove reviews with missing ratings, and truncate sequences to length 128.
The resulting split contains 1.9M training examples and 5{,}000 examples each for validation and test.

\textbf{BigPatent (long-form domain text).}
We use BigPatent \cite{sharma2019bigpatent} (patent abstracts), truncate sequences to length 128, and use 200K training examples with 5{,}000 examples each for validation and test.

\textbf{Control codes.}
We adopt control-code conditioning \cite{keskar2019ctrl}, a textual control code derived from structured attributes is prepended to each prompt and treated as part of the prefix.
For Yelp, the code includes business category and star rating (e.g., \texttt{"Business Type: Restaurant | Review Stars: 3.0"}).
For BigPatent, we prepend the CPC code.
This formatting is orthogonal to DP-OPD: we mask prompt tokens and compute distillation only on continuation tokens.

\subsection{Models}
\label{sec:models}

We follow the compression setting used by \textsc{DistilDP}: \textbf{GPT-2 Large} \cite{radford2019gpt2} is the teacher and \textbf{DistilGPT-2} \cite{sanh2019distilbert} is the student, yielding a $\sim$9.5$\times$ parameter gap (774M vs.\ 82M).
Unless otherwise noted, teacher and student share a tokenizer.

\subsection{Privacy and training configuration}
\label{sec:training_privacy}

We report results under a strong privacy regime: $\varepsilon = 2.0$ and $\delta = 1/N$, following standard practice for DP-SGD training \cite{abadi2016deep,dwork2006calibrating}.
DP-OPD's privacy guarantee is provided entirely by \emph{DP-SGD on student updates}; the teacher remains frozen and is not released.

Across datasets, we use learning rate $1\mathrm{e}{-4}$ and clipping norm $C=1.0$.
We train for 40 epochs on BigPatent and 5 epochs on Yelp.
Our best-performing configuration uses $\lambda=0.5$ and $\beta=0.5$.

\subsection{Baselines}
\label{sec:baselines}

\textbf{DP-SGD (student-only)} trains the student with DP-SGD and no teacher.
\textbf{DPKD} follows prior DP model compression formulations \cite{mireshghallah2022differentially} by applying DP-SGD to both teacher and student (splitting privacy budget).
\textbf{\textsc{DistilDP}} trains a DP teacher with DP-SGD, generates DP synthetic text (via post-processing), and trains the student on synthetic text using non-private optimization while distilling teacher outputs \cite{distildp}.

\subsection{Evaluation metric}
\label{sec:metric}

We use \textbf{test perplexity (PPL)} as the primary utility metric.
Lower PPL indicates better language modeling performance.

\subsection{Results}
\label{sec:main_results}

Table~\ref{tab:main_results} compares DP-OPD against baselines at $\varepsilon=2.0$.
Baseline values are taken from \textsc{DistilDP}, and we additionally report their DP-trained teacher (GPT-2 Large with DP-SGD at $\varepsilon=1.0$) for context, noting it is not directly comparable due to both model size and privacy budget.

\begin{table*}[t]
\centering
\small
\begin{tabular}{lccc}
\toprule
\textbf{Method} & $\boldsymbol{\varepsilon}$ & \textbf{Yelp PPL} $\downarrow$ & \textbf{BigPatent PPL} $\downarrow$ \\
\midrule
GPT-2 Large + DP-SGD (teacher) & 1.0 & 26.84 & 21.91 \\
\midrule
DistilGPT-2 + DP-SGD (student-only) & 2.0 & 48.12 & 41.80 \\
DistilGPT-2 + DPKD (off-policy DP KD) & 2.0 & 46.16 & 41.20 \\
DistilGPT-2 + \textsc{DistilDP} (DP synthetic text) \cite{distildp} & 2.0 & 44.15 & 32.43 \\
\midrule
\textbf{DP-OPD (ours)} & 2.0 & \textbf{41.68} & \textbf{30.63} \\
\bottomrule
\end{tabular}
\caption{Perplexity (PPL) comparison on Yelp and BigPatent datasets. Baseline numbers are reported in \textsc{DistilDP}\cite{distildp}.}
\label{tab:main_results}
\end{table*}

DP-OPD achieves the best student perplexity on both datasets at $\varepsilon=2.0$.
We attribute these gains to aligning the student with the teacher \emph{on the student’s own state distribution}, the prefixes the student actually visits at inference time, where DP-induced perturbations are most consequential.
Under DP-SGD, gradient noise can amplify exposure bias: the student is optimized on teacher-forced (off-policy) prefixes but must condition on its own imperfect generations at test time, so small local mistakes compound over long continuations.
DP-OPD mitigates this mismatch by querying the teacher on student-generated rollouts and applying dense token-level targets on exactly those visited states, yielding a more stable learning signal under strict privacy noise.

This perspective also helps explain why DP-OPD can outperform \textsc{DistilDP}.
Although \textsc{DistilDP} benefits from strong supervision via DP synthetic text, its distillation signal is computed on \emph{teacher-induced} trajectories, samples from the (DP) teacher under a particular decoding scheme, and finite synthetic set size, which can introduce an additional synthetic--student (and synthetic--real) distribution shift, especially for long-form generation.
In contrast, DP-OPD continually corrects the student \emph{in-distribution} by distilling on student rollouts conditioned on real prompts, reducing compounding errors without requiring DP optimization of the teacher or an offline synthetic generation stage.

\subsection{Compute and practical cost}
\label{sec:compute}

Beyond utility, DP-OPD simplifies private compression and reduces the practical resource footprint relative to synthesis-based approaches.
\textsc{DistilDP} requires (i) DP-SGD training of a large teacher, (ii) large-scale synthetic text generation, and (iii) separate student training on synthetic data, and reports using large multi-GPU configurations across these stages (e.g., 8$\times$40GB A100 for DP training components and 8 GPUs for synthetic-text student training), along with multi-day end-to-end runtimes. \cite{distildp}

In contrast, DP-OPD performs distillation online during a single DP-SGD training run of the student and eliminates offline synthetic dataset generation.
All of our experiments were conducted on a \textbf{single RTX A6000 (40GB)}.
For BigPatent, training took \textbf{2 days 7 hours}.
For Yelp, training took \textbf{3 days 16 hours} for \textbf{5 epochs}.

While DP-OPD introduces rollout overhead due to on-policy generation, in practice $\lambda=0.5$ provides a favorable tradeoff between throughput and on-policy alignment, reducing rollout frequency while maintaining the best observed perplexity.
We emphasize that these numbers reflect different hardware and implementation choices and are not a fully normalized FLOP comparison; rather, they indicate that DP-OPD can achieve competitive (and in our case improved) private compression performance with a smaller and operationally simpler compute footprint.

\subsection{Ablation studies}
\label{sec:ablations}

\begin{figure}[t]
  \centering
  \includegraphics[width=1.0\linewidth]{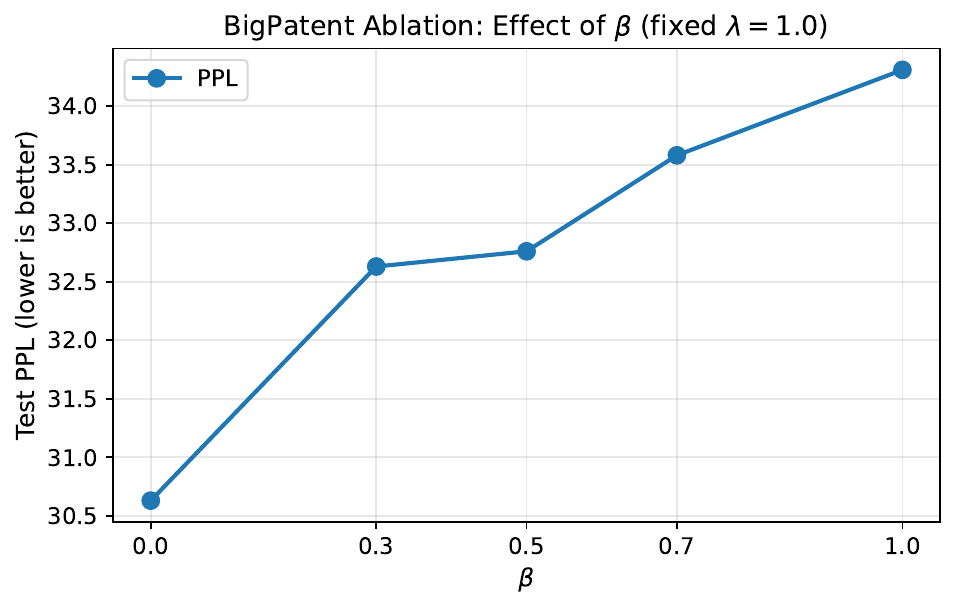}
  \caption{Ablation on BigPatent: test perplexity as a function of the GKD divergence parameter $\beta$ with $\lambda=1.0$ (on-policy every step). Lower $\beta$ yields lower PPL in our setting, with the best performance at $\beta=0$.}
  \label{fig:beta_sweep}
\end{figure}

\textbf{Effect of $\beta$ (GKD divergence).}
We fix $\lambda=1.0$ and sweep $\beta \in \{0, 0.3, 0.5, 0.7, 1.0\}$ on BigPatent, reporting test perplexity as a function of $\beta$
(Figure~\ref{fig:beta_sweep}).
All settings are trained for 40 epochs.

Overall, smaller $\beta$ yields better test perplexity: $\beta=0$ achieves the best PPL (30.63), and PPL increases as $\beta$ grows (32.63 at $\beta=0.3$, 32.76 at $\beta=0.5$, 33.58 at $\beta=0.7$, and 34.31 at $\beta=1.0$).
This trend is consistent with the role of $\beta$ in generalized distillation objectives~\cite{agarwal2024opd}: lower $\beta$ corresponds to a more forward-KL-like alignment that encourages the student to cover the teacher's predictive distribution, which is closely tied to next-token likelihood and thus perplexity.
As $\beta$ increases toward a more reverse-KL-like regime, the objective becomes more mode-seeking (emphasizing high-probability teacher modes rather than full support coverage), which can be beneficial for certain generation properties but does not necessarily optimize perplexity, matching the observed PPL degradation.

% \textbf{Effect of $\lambda$ (on-policy ratio).}
% Fix $\beta=1.0$ and sweep $\lambda \in \{0, 0.3, 0.5, 0.7, 1.0\}$; report PPL vs.\ $\lambda$ and throughput (tokens/sec).

% \begin{figure}[t]
%   \centering
%   % \includegraphics[width=0.8\linewidth]{lambda_sweep.pdf}
%   \caption{(Placeholder) BigPatent test PPL vs.\ $\lambda$ with $\beta=0.5$.}
%   \label{fig:lambda_sweep}
% \end{figure}

\subsection{Limitations}
\label{sec:limitations}

DP-OPD trades offline synthetic generation for \emph{online} student rollouts and teacher evaluation on those rollouts.
For long continuation lengths or larger students, rollout generation can dominate wall-clock time, and frequent teacher queries can increase training-time inference cost.
This is a different cost profile than synthesis-based pipelines (e.g., \textsc{DistilDP}), which can amortize teacher access by generating a fixed synthetic corpus once and then training the student without further teacher calls.

The method also assumes access to a frozen teacher during training; if teacher queries are rate-limited or costly, the bottleneck may shift from training compute to teacher inference budget.
Finally, although we follow prior work in treating control codes as non-sensitive conditioning signals, in some deployments the metadata distribution itself may be sensitive; handling this rigorously requires an end-to-end threat model and may require additional privatization of metadata.

\section{Conclusion}
\label{sec:conclusion}

We introduced \textbf{DP-OPD}, a synthesis-free framework for differentially private compression of autoregressive language models that combines on-policy distillation with standard DP-SGD accounting. DP-OPD trains the student on \emph{self-generated} trajectories and uses a frozen teacher to provide dense token-level targets on the states the student actually visits, while enforcing privacy solely through DP-SGD on the student updates. This directly addresses train--test mismatch and exposure bias, which can be exacerbated by DP noise.

Under a strict privacy budget ($\varepsilon=2.0$), DP-OPD improves perplexity over DP-SGD fine-tuning and off-policy DP distillation, and  outperforms synthesis-based DP distillation on both Yelp and BigPatent. Beyond utility, DP-OPD simplifies private compression by collapsing the pipeline into a single DP student-training loop, avoiding DP training of the teacher and offline synthetic text generation.

Future work includes characterizing the compute--utility tradeoff as rollout frequency and length scale, testing it with tokenizer-mismatched teacher--student pairs and constrained teacher-query settings, and refining threat models where conditioning metadata (control codes) is sensitive.

\bibliography{manuscript_paper}

@inproceedings{distildp,
  title     = {Differentially Private Knowledge Distillation via Synthetic Text Generation},
  author    = {Flemings, James and Annavaram, Murali},
  booktitle = {Findings of the Association for Computational Linguistics: ACL 2024},
  year      = {2024},
  doi       = {10.18653/v1/2024.findings-acl.769},
  url       = {https://aclanthology.org/2024.findings-acl.769/}
}

@article{agarwal2024opd,
  title   = {On-Policy Distillation of Language Models: Learning from Self-Generated Mistakes},
  author  = {Agarwal, Rishabh and Vieillard, Nino and Zhou, Yongchao and Stanczyk, Piotr and Ramos, Sabela and Geist, Matthieu and Bachem, Olivier},
  journal = {arXiv preprint arXiv:2306.13649},
  year    = {2023},
  note    = {To appear / version cited as 2024 in some venues}
}

@inproceedings{yue2023synthetic,
  title={Synthetic text generation with differential privacy: A simple and practical recipe},
  author={Yue, Xiang and Inan, Huseyin and Li, Xuechen and Kumar, Girish and McAnallen, Julia and Shajari, Hoda and Sun, Huan and Levitan, David and Sim, Robert},
  booktitle={Proceedings of the 61st Annual Meeting of the Association for Computational Linguistics (Volume 1: Long Papers)},
  pages={1321--1342},
  year={2023}
}

@inproceedings{sanh2019distilbert,
  title     = {DistilBERT, a distilled version of BERT: smaller, faster, cheaper and lighter},
  author    = {Sanh, Victor and Debut, Lysandre and Chaumond, Julien and Wolf, Thomas},
  booktitle = {NeurIPS EMC\textsuperscript{2} Workshop},
  year      = {2019}
}

@article{keskar2019ctrl,
  title={Ctrl: A conditional transformer language model for controllable generation},
  author={Keskar, Nitish Shirish and McCann, Bryan and Varshney, Lav R and Xiong, Caiming and Socher, Richard},
  journal={arXiv preprint arXiv:1909.05858},
  year={2019}
}

@misc{opd_h4_space,
  title        = {Unlocking On-Policy Distillation for Any Model Family},
  author       = {{Hugging Face H4}},
  year         = {2025},
  month        = oct,
  howpublished = {\url{https://huggingface.co/spaces/HuggingFaceH4/on-policy-distillation}},
  note         = {Accessed 2026-01-31}
}

@inproceedings{abadi2016deep,
  title     = {Deep Learning with Differential Privacy},
  author    = {Abadi, Mart{\'i}n and Chu, Andy and Goodfellow, Ian and McMahan, H. Brendan and Mironov, Ilya and Talwar, Kunal and Zhang, Li},
  booktitle = {Proceedings of the 2016 ACM SIGSAC Conference on Computer and Communications Security (CCS)},
  year      = {2016},
  doi       = {10.1145/2976749.2978318}
}

@inproceedings{dwork2006calibrating,
  title     = {Calibrating Noise to Sensitivity in Private Data Analysis},
  author    = {Dwork, Cynthia and McSherry, Frank and Nissim, Kobbi and Smith, Adam},
  booktitle = {Theory of Cryptography Conference (TCC)},
  year      = {2006},
  doi       = {10.1007/11681878_14}
}

@inproceedings{bengio2015scheduled,
  title     = {Scheduled Sampling for Sequence Prediction with Recurrent Neural Networks},
  author    = {Bengio, Samy and Vinyals, Oriol and Jaitly, Navdeep and Shazeer, Noam},
  booktitle = {Advances in Neural Information Processing Systems (NeurIPS)},
  year      = {2015}
}

@article{ranzato2016sequence,
  title   = {Sequence Level Training with Recurrent Neural Networks},
  author  = {Ranzato, Marc'Aurelio and Chopra, Sumit and Auli, Michael and Zaremba, Wojciech},
  journal = {arXiv preprint arXiv:1511.06732},
  year    = {2015}
}

@article{li2021large,
  title   = {Large Language Models Can Be Strong Differentially Private Learners},
  author  = {Li, Xuechen and Tram{\`e}r, Florian and Liang, Percy and Hashimoto, Tatsunori},
  journal = {arXiv preprint arXiv:2110.05679},
  year    = {2021}
}

@article{papernot2016pate,
  title   = {Semi-supervised Knowledge Transfer for Deep Learning from Private Training Data},
  author  = {Papernot, Nicolas and Abadi, Mart{\'i}n and Erlingsson, {\'U}lfar and Goodfellow, Ian and Talwar, Kunal},
  journal = {arXiv preprint arXiv:1610.05755},
  year    = {2016}
}

@article{majmudar2022dpdecoding,
  title   = {Differentially Private Decoding in Large Language Models},
  author  = {Majmudar, Jimit and Dupuy, Christophe and Peris, Charith and Smaili, Sami and Gupta, Rahul and Zemel, Richard},
  journal = {arXiv preprint arXiv:2205.13621},
  year    = {2022}
}

@article{mattern2022dplm,
  title   = {Differentially Private Language Models for Secure Data Sharing},
  author  = {Mattern, Justus and Jin, Zhijing and Weggenmann, Benjamin and Sch{\"o}lkopf, Bernhard and Sachan, Mrinmaya},
  journal = {arXiv preprint arXiv:2210.13918},
  year    = {2022}
}

@article{kurakin2023privateSynth,
  title   = {Harnessing Large-Language Models to Generate Private Synthetic Text},
  author  = {Kurakin, Alexey and Ponomareva, Natalia and Syed, Umar and MacDermed, Liam and Terzis, Andreas},
  journal = {arXiv preprint arXiv:2306.01684},
  year    = {2023}
}

@article{mireshghallah2022differentially,
  title={Differentially private model compression},
  author={Mireshghallah, Fatemehsadat and Backurs, Arturs and Inan, Huseyin A and Wutschitz, Lukas and Kulkarni, Janardhan},
  journal={Advances in Neural Information Processing Systems},
  volume={35},
  pages={29468--29483},
  year={2022}
}

@inproceedings{sharma2019bigpatent,
  title     = {BigPatent: A Large-Scale Dataset for Abstractive and Coherent Summarization},
  author    = {Sharma, Eva and Li, Chen and Zhang, Lu Wang},
  booktitle = {Proceedings of the 57th Annual Meeting of the Association for Computational Linguistics (ACL)},
  year      = {2019}
}

@article{radford2019gpt2,
  title        = {Language Models are Unsupervised Multitask Learners},
  author       = {Radford, Alec and Wu, Jeffrey and Child, Rewon and Luan, David and Amodei, Dario and Sutskever, Ilya},
  journal      = {OpenAI Technical Report},
  year         = {2019},
  howpublished = {\url{https://cdn.openai.com/better-language-models/language_models_are_unsupervised_multitask_learners.pdf}}
}
\bibliographystyle{icml2026}

%%%%%%%%%%%%%%%%%%%%%%%%%%%%%%%%%%%%%%%%%%%%%%%%%%%%%%%%%%%%%%%%%%%%%%%%%%%%%%%
%%%%%%%%%%%%%%%%%%%%%%%%%%%%%%%%%%%%%%%%%%%%%%%%%%%%%%%%%%%%%%%%%%%%%%%%%%%%%%%
% APPENDIX
%%%%%%%%%%%%%%%%%%%%%%%%%%%%%%%%%%%%%%%%%%%%%%%%%%%%%%%%%%%%%%%%%%%%%%%%%%%%%%%
%%%%%%%%%%%%%%%%%%%%%%%%%%%%%%%%%%%%%%%%%%%%%%%%%%%%%%%%%%%%%%%%%%%%%%%%%%%%%%%
\newpage
\appendix
\onecolumn


% You can have as much text here as you want. The main body must be at most $8$
% pages long. For the final version, one more page can be added. If you want, you
% can use an appendix like this one.

% The $\mathtt{\backslash onecolumn}$ command above can be kept in place if you
% prefer a one-column appendix, or can be removed if you prefer a two-column
% appendix.  Apart from this possible change, the style (font size, spacing,
% margins, page numbering, etc.) should be kept the same as the main body.
%%%%%%%%%%%%%%%%%%%%%%%%%%%%%%%%%%%%%%%%%%%%%%%%%%%%%%%%%%%%%%%%%%%%%%%%%%%%%%%
%%%%%%%%%%%%%%%%%%%%%%%%%%%%%%%%%%%%%%%%%%%%%%%%%%%%%%%%%%%%%%%%%%%%%%%%%%%%%%%

\end{document}